\documentclass[conference]{IEEEtran}
\ifCLASSINFOpdf
	\usepackage[pdftex]{graphicx}
	\graphicspath{{figures/}}
	\DeclareGraphicsExtensions{.pdf,.jpeg,.png}
\else
	\usepackage[dvips]{graphicx}
	\graphicspath{{figures/}}
	\DeclareGraphicsExtensions{.eps}
\fi
\PassOptionsToPackage{hyphens}{url}
\usepackage{hyperref}
\usepackage[autostyle=true]{csquotes}  

\usepackage{eso-pic}
\usepackage{lipsum}
\AddToShipoutPicture{%
     \AtTextLowerLeft{%
         \put(-5,-30){\fbox{Accepted for presentation at the 14th IEEE International Conference on Semantic Computing -- Resource Track}}%
     }%
}

\let\labelindent\relax
\usepackage{enumitem}
\usepackage{tabularx}

\usepackage[textsize=tiny,
]{todonotes}

\usepackage{tikz}
\usepackage{pgfplots}
\usepackage{tikz-qtree}
\usepackage{algorithm}
\usepackage{algpseudocode}

\makeatletter
\renewcommand*\ALG@name{Script:}
\makeatother

\begin{document}
%
\title{At Your Command!\\ An Empirical Study on How Laypersons\\ Teach Robots New Functions}

\author{\IEEEauthorblockN{Sebastian Weigelt, Vanessa Steurer, Walter F. Tichy}
	\IEEEauthorblockA{Karlsruhe Institute of Technology\\Institute for Program Structures and Data Organization\\
		Karlsruhe, Germany\\
		\href{mailto:weigelt@kit.edu}{weigelt@kit.edu}, \href{mailto:vanessa.steurer@web.de} {vanessa.steurer@web.de}, \href{mailto:tichy@kit.edu}{tichy@kit.edu}}
}



\maketitle


\begin{abstract}
	Even though intelligent systems such as Siri or Google Assistant are enjoyable (and useful) dialog partners, users can only access predefined functionality.
	Enabling end-users to extend the functionality of intelligent systems will be the next big thing.
	To promote research in this area we carried out an empirical study on how laypersons teach robots new functions by means of natural language instructions.
	The result is a labeled corpus consisting of 3168 submissions given by 870 subjects.

	The analysis of the dataset revealed that many participants used certain wordings to express their wish to teach new functionality;
	two corresponding trigrams are among the most frequent.
	On the contrary, more than one third (36.93\%) did not verbalize the teaching intent at all.
	We labeled the semantic constituents in the utterances: declaration (including the name of the function) and intermediate steps.
	The full corpus is publicly available: \url{http://dx.doi.org/10.21227/zecn-6c61}
\end{abstract}

\section{Introduction}

Over the last few years, Conversational Interfaces (CI) have become more and more popular.
With the rise of a new generation of intelligent systems, CIs became established alongside graphical user interfaces as a type of human computer interaction.
Today, users naturally communicate with intelligent systems, such as virtual assistants or household robots~\cite{christensenCognitive2010,bellegardaSpoken2014a}. 
For instance, one can easily arrange an appointment by talking to Apple's Siri or Google Assistant.
However, the technical advancement of CIs seems to fall behind their increasing presence.
The usability of CIs has to improve soon to meet the users' expectations.
On the one hand, users will expect that systems understand increasingly complex queries and actually grasp the user's intent rather than reacting to predefined phrases\footnote{We have considered this matter in several publications. See e.g. \cite{weigeltContext2017,weigeltDetection2018,weigeltDetection2018a}, and \cite{weigeltUnsupervised2019} for discussions of several sub-tasks regarding intent extraction.}.
On the other hand, users will not much longer accept that they can only access built-in functionality with CIs.
For the time being, there is no way to add new functions to intelligent systems through CIs.
Instead, only the developers of CIs can extend their functionality.
However, it would be much more efficient to enable the user to enhance the CI through the CI itself.
This would engender a new kind of end-user development.
The reason why this option is virtually unexplored is the lack of empirical foundation.
Even though, corpora for CIs have been created in recent years~\cite{bastianelliHuRIC2014,salesSemEval20172017}, there is no dataset to be found that focuses on function extensions.

To fill this gap and promote research in this area we have created a labeled corpus that examines how laypersons teach intelligent systems new functions by means of natural language instructions.
The corpus gives insight on the language and structure laypersons use in this situation\footnote{Our research focuses on the description of functions (in the form of actions). Thus, we do \emph{not} consider other types of describable knowledge, such as common-sense or environmental knowledge.}.
To be more exact, the corpus is supposed to answer the following research questions (among others):
\begin{enumerate}
	\item Do laypersons always clearly state that they wish to add new functionality?
	\item Can the wish for extension (and the name of the new function) be clearly separated from the actions that are to be performed?
\end{enumerate}

To gather the natural language instructions, we have created a study comprising a set of four scenarios.
All scenarios consider instructions for a humanoid household robot in a kitchen environment.
We chose this setting, since humans are familiar with tasks (and task descriptions) in this domain.
Thus, we expected naturally phrased utterances.
Subjects were introduced to the setting and the task to give the robot instructions to learn a new function.
We used the micro-tasking platform \emph{Prolific} for our study.
We collected 3470 written task descriptions (3168 after an in-depth sanity check) from 870 subjects.
An exemplary task description given by a subject is:

\enquote{\emph{preparing a cup of coffee means you have to put a coffee mug under the dispenser and then press the red button on the coffee machine thats how you make some coffee}}

We attach two types of labels to the descriptions.
First, we attach binary labels to entire descriptions which denote the presence of an explicitly stated wish for function extension, i.e. a teaching intent.
That means, we label whether a description includes a declaration of new functionality, including a name for the new function, or instead, solely a sequence of actions.
Second, we attach per-word labels that describe the semantic function of the resp. word regarding function extensions.
Therefore, we distinguish words which are part of the wish for extension and function naming, the sequence of actions to be performed, and words without valuable information (in terms of our objective).
The full corpus, including raw data, labeled data, meta-data, and scenario descriptions, is publicly available: \url{http://dx.doi.org/10.21227/zecn-6c61}

\section{Related Work}
\label{sec:rw}

The vision of controlling and programming computer systems with natural language goes back to Jane Sammet in 1966~\cite{sammetUse1966}.
Since then, the development of such systems proceeded slowly and renowned researcher doubted its practicability~\cite{hillWouldn1972,dijkstrafoolishness1979}.
Until today, this research area lacks profound theoretical foundations and -- to a great extent -- an empirical basis.
Subsequently, we present studies on natural language programming
, approaches for teaching new functionality with natural language, and existing public corpora.

\paragraph{Studies}
In 1981, Miller conducted a study regarding the general feasibility of natural language programming~\cite{millerNatural1981}.
For the study fourteen college students dictated instruction sequences, that an intelligent system could have executed.
The author describes a set of difficult to clear obstacles he recognized, such as semantic and context, which can hardly be interpreted correctly in all cases, and the lack of world knowledge.
Despite these bleak prospects researchers such as Nielsen postulate that future computer interfaces should speak the user's language and view interactions from a user's perspective~\cite{nielsenUsability1993}.
Biermann et al. confirm another of Miller's observation: humans tend to describe procedures step-by-step, which suggests that they might be transferable to actual programs~\cite{biermannExperimental1983}.
Pane et al. conducted two studies in which subjects (36 fifth graders and nineteen adults in sum) were supposed to give descriptions of the computer game \emph{Pac Man} in natural language~\cite{paneStudying2001}.
One of the many interesting findings is, that humans often describe things implicitly, e.g. looping: \enquote{until} or \enquote{and so on.}
However, the authors conclusively state that, \enquote{[...] natural language solutions are satisfactory, but are different in style [...]}
Based on the study by Pane et. al, Lieberman et al. investigated, how certain natural language structures can be translated to source code~\cite{liebermanFeasibility2006}.
From their observations they derived rules such as \enquote{nouns refer to objects} and \enquote{verbs are functions}.

\paragraph{Approaches for Teaching New Function with Natural Language}
Present approaches for teaching new function with natural language do not use extensive datasets.
Instead, most of them rely on heuristics.
\emph{SmartSynth} by Le et al. allows laypersons to interactively create short scripts for mobile phones~\cite{leSmartSynth2013}.
The approach synthesizes API calls employing linguistic features and incorporates domain knowledge for type checking.
She et al. teach robots new actions through natural language instructions~\cite{sheTeaching2014}.
Besides a specialized semantic processor, they incorporate domain knowledge and the robot's perception in the process.
Markievicz et al. use heuristics based on semantic roles to synthesize executable code for robotic systems from natural instructions originally written for humans~\cite{markieviczReading2017}.
Other approaches integrate linguistic resources into other techniques of end-user development.
Manshadi et al. use unconstrained natural language instructions to improve their classifier for programming by example (based on directed acyclic graphs)~\cite{manshadiIntegrating2013}.
Li et al. consider natural language as one kind of input to their multi-modal-approach \emph{APPINITE}~\cite{liAPPINITE2018}.
With APPINITE a user can define action sequences by selecting actions, recording touch sequences, and giving natural language descriptions.

\paragraph{corpora}
Among the vast amount of linguistic corpora only a few labeled corpora for programming in natural language exist.
Tellex et al. created a corpus for their approach on \emph{Understanding Natural Language Commands for Robotic Navigation and Mobile Manipulation}~\cite{tellexUnderstanding2011}.
They presented videos of a robotic forklift to subjects and let them describe what they see (as instructions).
To gather the data, Tellex et al. used the microtasking plattform \emph{Amazon Mechanical Turk}\footnote{Amazon Mechanical Turk: \url{https://www.mturk.com/}} and were able to collect instruction sequences from 45 subjects with thirteen instructions on average.
In 2014, Bastianelli et al. released the \emph{Human Robot Interaction Corpus} (HuRIC)~\cite{bastianelliHuRIC2014}.
It comprises three datasets from the robotic domain with 570 audio recordings taken from 51 subjects in sum.
Besides a manual transcript, the corpus includes the following labels for each instance:
lemmas, part-of-speech tags, and dependency trees.
The SemEval-2017 Task 11, proposed by Sales et al., comes with a labeled corpus for end-user development using natural language~\cite{salesSemEval20172017}.
However, the focus is on single API calls.
Additionally, it makes the assumption that domain knowledge is present (and properly modeled), e.g. about the functions of a mobile phone, including method calls, parameters, etc.

\paragraph{Summary and Positioning}
Although there are studies on programming in natural language, to the best of our knowledge none of them considered teaching new functionality; the same applies to publicly accessible corpora.
Regarding systems for programming in natural languages, there are a few that address defining new functionality.
However, they lack an extensive data basis (none is described or accessible at least).

\section{Study}
\label{sec:data}

This section details the study.
Subsequently, we first present design principles and research questions for our study (\autoref{sec:data:subsec:design}).
Then, we give details about the used scenarios (\autoref{sec:data:subsec:scenarios}) and the actual data collecting (\autoref{sec:data:subsec:conduction}) before we discuss the labeling process (\autoref{sec:data:subsec:labeling}).
Finally, we analyze the results, including meta-data, subject characteristics, and an in-depth inspection of the gathered material(\autoref{sec:data:subsec:analysis}).

\subsection{Research Questions \& Design Principles}
\label{sec:data:subsec:design}

With our study we want to examine the language and structure used by laypersons when describing new functionality to a subject, e.g. another human or an intelligent system such as a humanoid robot.
Proceeding from this rather abstract objective, we collected a set of research questions.
From this set we chose the four following we hope to answer with the study:

\begin{itemize}[align=right, itemindent=0.6cm]
	\item  [\emph{RQ1:}] Do laypersons always clearly state that they wish to add new functionality?
	\item [\emph{RQ2:}] Can the wish for extension (and the name of the new function) be clearly separated from the actions that are to be performed?
	\item [\emph{RQ3:}] Do laypersons tend to add non-descriptive or meaningless statements and if so, to which extent?
	\item [\emph{RQ4:}] Do laypersons repeatedly use particular phrases to declare certain intentions, e.g. the wish for extension?
\end{itemize}

Beside these research questions we defined the following objectives for our study.
First, we do not restrict the subjects' language.
They are supposed to speak as they like.
Of course this entails plenty of challenges, e.g. a wide range of different (and potentially complex) phrases and ambiguity.
Second, we collect spontaneous speech.
Again, this entails issues, such as incorrect wording or grammar mistakes.
Finally, we aim to create a corpus that focuses on the research topic rather than the actual domains.
Thus, we have to create scenarios that are tangible on the one hand and generally valid on the other.

\subsection{Scenarios}
\label{sec:data:subsec:scenarios}

We developed four scenarios for our study.
In each scenario, subjects are supposed to teach a humanoid household robot new skills.
Therefore, they should imagine that the robot is like a child with knowledge about its environment and basic abilities.
All scenarios are set-up in a kitchen environment.
As vivid example for a robot we chose the \emph{ARMAR-III}\footnote{ARMAR-III: \url{https://his.anthropomatik.kit.edu/english/241.php}}~\cite{asfourARMARIII2006}.

\begin{figure}[t]
	\begin{center}
		\includegraphics[width=\columnwidth]{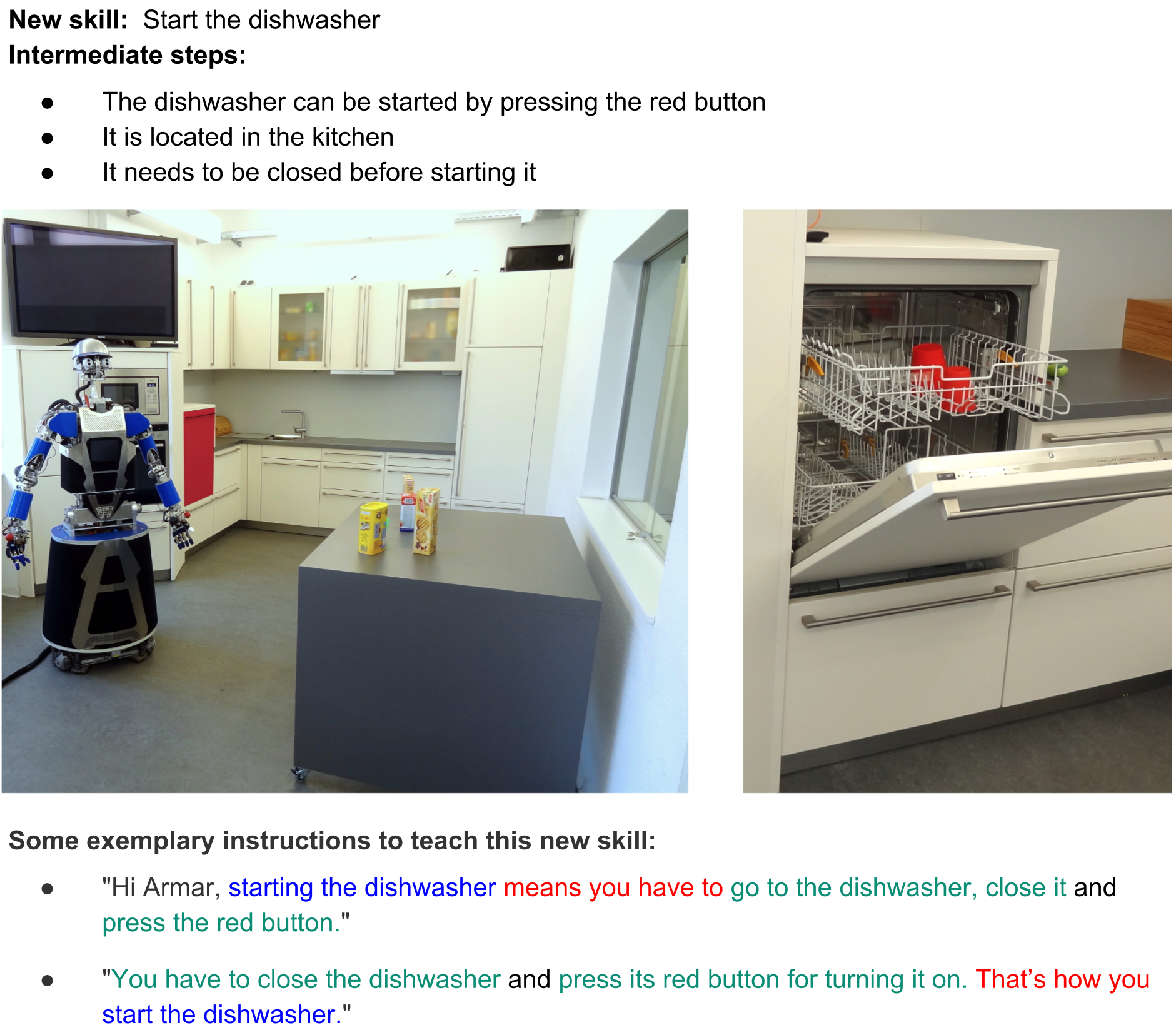}
	\end{center}
	\caption{The example scenario from the study including solutions.}
	\label{fig:scen3}
\end{figure}

The scenarios are \emph{greeting someone}, \emph{preparing coffee}, \emph{serving drinks}, \emph{and setting a table for two}.
Each contains a short name (of the skill to be taught), a list of possible intermediate steps, and pictures depicting the setting.
We created a fifth scenario (\emph{starting the dishwasher}) that also includes example solutions.
This additional scenario is used to familiarize the participants with the setting and to present potential valid solutions.
\autoref{fig:scen3} depicts the fifth scenario; it considers the skill \emph{start the dishwasher}.
The examples emphasize the components of a valid response: a name for the new skill (in blue), the explicit expression of the wish to teach something (in red) and intermediate steps.

We decided not to use videos like Tellex et al.~\cite{tellexUnderstanding2011} since we thought that they might distract the subjects.
Also, we wanted the subjects to decide which intermediate steps are needed to learn the skill.
All descriptions are short and free of technical terms.
We also used simple phrasings and limited vocabulary\footnote{This approach was first described by Miller~\cite{millerNatural1981} and implemented consequently in the study by Pane et al.~\cite{paneStudying2001}}. 
We introduce the subjects with an explanatory scenario (including valid answers) at the risk of affecting the subjects' responses since we feared that the task might be too demanding otherwise\footnote{Campagna et al. observed, that subjects phrase queries differently even if they are provided with examples~\cite{campagnaAlmond2017}.}.
In the introductory part of the study we instruct subjects to use different wording for each scenario, not to parrot the examples, and speak as freely and spontaneously as possible.
Furthermore, the subjects should imagine the robot stands in front of them while they speak.
We also introduced the subjects to the objective of the study and pointed to the need to explicitly indicate that they want to teach a new skill.
Additionally, we reminded the subjects that a new skill is most likely composed of intermediate steps.
The scenario descriptions used for the study can be found on \url{https://forms.gle/Y9XX3g3LxZofZHYL9}

\subsection{Data Collection}
\label{sec:data:subsec:conduction}

Each subject participates in a single session that is composed of an introduction, the four scenarios  (as detailed in \autoref{sec:data:subsec:scenarios}), and a short questionnaire.
There is a short task description for each scenario.
The subjects are supposed to respond to each task; we call this response \emph{a submission}.

We used \emph{Prolific}\footnote{Prolific: \url{https://www.prolific.co/}} to collect data.
Prolific is an online micro-tasking platform that focuses on research tasks, e.g. user studies.
With over 70,000 registered users\footnote{As of 09/06/2019}, one easily recruits a sufficient number of participants for a study.
We decided not to conduct a stationary study at our institute, which we have also done in the past.
Even tough we give up some control over the study in an online setting, we would have never been able to recruit enough participants with a stationary study.
Since our study is of exploratory nature in the first place, we decided that magnitude matters more than full control.
For the same reason, we did not specify any eligibility requirements.

Another decision we had to make concerned the mode of input.
Originally, we planed to take audio recordings.
However, from studies past we have learned that recordings from online studies are of poor quality;
most subjects use poor microphones.
The consequence of low quality recordings is a cumbersome transcription process; automatic transcription becomes inconceivable.
Additionally, we think that the need for a microphone imposes another hurdle to overcome for potential participants, which might result in a lower willingness to participate.
Therefore, we preferred textual submissions for the study.
However, we explicitly pointed out that the participants should imagine a situation in which they literally talk to the robot and respond spontaneously.

 \begin{table}[t]
	\caption{Overview of the submissions from the test run and the full study.}
	\label{tab:overview}
	\begin{center}
		\begin{tabular}{lrrr}
			& test run & study & sum \\
			\hline
			participants & 10 & 860 & 870 \\
			submissions & 30 & 3440 & 3470 \\
			accepted submissions & 30 & 3138 & 3168 \\
		\end{tabular}
	\end{center}
\end{table}

The submissions are not constrained in length\footnote{Note that both Miller~\cite{millerNatural1981} and Biermann et al.~\cite{biermannExperimental1983} limited the submissions in length.
However, our first priority for this study is to let the subjects speak freely, which implicates unrestricted length in our view.}. 
We decided to define a time limit of 25 minutes per session, to retain internal validity.
The estimated completion time is seven minutes.
Each subject is allowed to participate in the study only once.
However, they may abort the study at any time.

A session starts with an explanatory page.
We introduce the subjects to the study objectives and boundary conditions.
Then, after presenting the exemplary scenario, all four scenarios are presented to the subjects one after another.
Subjects can leave their responses in a text box beneath the scenario description.
On a final page we ask all subjects to give us information about their programming skills and whether they are native English speakers or not.

To verify our study design we performed a test run with ten participants and three scenarios.
We put the study online on November 5\textsuperscript{th} 2018, 18:10 CET.
In less than 20 minutes ten subjects participated in the study.
The quality of the submissions were better than we anticipated.
Also, there were no major flaws or ambiguities in the study design.
However, we decided to add a fourth scenario to increase diversity.

After we had designed the final scenario, we ran the main study.
This time we used four scenarios and allowed 860 subjects to participate.
We released it on December 7\textsuperscript{th} 2018, 15:25 CET.
In less than three days we collected 3440 submissions.
The process at Prolific involves a manual review of the submissions of all participants.
We only conducted a simple sanity check during the run time of the study:
\begin{enumerate}
	\item Is every text field filled?
	\item Does a submission obviously contains nonsense?
\end{enumerate}

This manual approvement process accounted for the main time.
For each rejected submission another subject participated.
After the completion of the study, we carried out an in-depth review of the submissions.
We eliminated all submissions that clearly did not meet the objective of the study.
The range of eliminated submissions reaches from participants that simply misunderstood the task 
to those that intentionally submitted nonsense, such as definitions from Wikipedia, short stories, or even poems.
We had to decline 302 submissions, which accounts for 8.78\%.
\autoref{tab:overview} summarizes the numbers for the test run and the study.

\subsection{Labeling}
\label{sec:data:subsec:labeling}

After we eliminated all inappropriate submissions, we labeled all remaining submissions.
A first, rough analysis of the dataset revealed that about one third of all submissions are sequences of actions rather than descriptions of new functionality.
Although we distinctly pointed out in the introductory part of the study that each submission must explicitly state that a new function is to be learned, many subjects missed out either the wish for extension or a name for the new function (or both) and described the intermediate steps only.
Therefore, we introduced the binary label set:
\begin{itemize}
	\item \emph{Function Description}: used for submissions that contain a teaching intent, i.e. an explicitly stated wish for extension and a name for the new function. 
	\item \emph{Instruction Sequence}: used for submissions that describe a sequence of actions only, i.e. all other submissions.
\end{itemize}
Note that these binary labels are on a per-submission level.
With the help of these labels, we will be able to answer \emph{RQ1} in \autoref{sec:data:subsec:analysis}.
\begin{figure}
	\setlength{\tabcolsep}{1pt}
	\begin{tabularx}{\columnwidth}{lX}				
		E1: & [ \emph{armar coffee is a beverage that people like to drink} ]\textsubscript{\textbf{MISC}} [ \emph{in order to make coffee} ]\textsubscript{\textbf{DECL}} [ \emph{you have to locate the cups next to the machine put one cup under the dispenser and lastly press the red button on the coffee machine} ]\textsubscript{\textbf{SPEC}}\\
		E2: & [ \emph{to greet a person} ]\textsubscript{\textbf{DECL}} [ \emph{look into the persons eyes wave your robot hand and say hello} ]\textsubscript{\textbf{SPEC}} [ \emph{this is how you greet someone} ]\textsubscript{\textbf{DECL}}
		
	\end{tabularx}
	\caption{Two examples depicting the labels \emph{Declaration} (DECL), \emph{Specification} (SPEC), and \emph{Miscellaneous} (MISC). We used a block-like annotation scheme for the sake of readability.}
	\label{fig:example-labels}
\end{figure}
For analyzing the submissions that contain a function description we defined the following labels\footnote{Note that we only consider submissions with \emph{Function Description}-labels.}:
\begin{itemize}
	\item \emph{Declaration}: This label is attached to all parts of a submission that contain either the wish for extension or the name of the new function.
	We decided to consolidate these two parts in one label for two reasons.
	First, we observed that a clear separation is virtually impossible for many submissions.
	Second, we believe that these two parts are semantically related, which makes a shared label reasonable. Example: \enquote{in order to make coffee}.
	\item \emph{Specification}: This label is attached to all parts of a submission that describe the intermediate steps of the function to be learned. Example:
	\enquote{go to the sink}
	\item \emph{Miscellaneous}: This label is attached to all parts that do not 
	fit in the above categories, i.e. they are irrelevant in our context. Example: \enquote{hey robot}.
\end{itemize}

We will be able to answer \emph{RQ2} with the first two labels.
With the help of the \emph{Miscellaneous}-label we will answer \emph{RQ3}.
\autoref{fig:example-labels} shows two labeled examples.
The first example comprises a part without useful information (\emph{MISC}-label).
The second has two declarative parts (\emph{DECL}-label).
The second declaration resembles the first but uses a different wording.

Note that we use per-word labels instead of block labels (as denoted in the example).
We do so since we observed that participants tend to mix different parts in their submissions.
Sometimes even \emph{Declaration} and \emph{Specification} are hard to separate.
See \autoref{sec:data:subsec:analysis} for a thorough discussion.

The dataset was labeled by the first and the second author.
We decided not to label separately, since we experienced that almost half of the labels were arguable.
Therefore, we changed over to a joint labeling process.
We first attached the binary labels.
We decided to follow a rather strict approach, i.e. in case of doubt, we attached the \emph{Instruction Sequence}-label.
We first tried to determine the declarative parts for the second label set.
For the remaining words we tried to separate \emph{Specification} and all parts that do not carry any relevant information, e.g. greetings and common-sense teachings.
Here we decided to rather mark words as relevant, i.e. attaching the labels \emph{Declaration} or \emph{Specification}, then irrelevant (label \emph{Miscellaneous}).
During the second labeling stage, we were able to determine some false positives from the first stage, since we were unable to find a declarative part.
We corrected all these false labels.
The results of the labeling process will be discussed in \autoref{sec:data:subsec:analysis}.

\subsection{Analysis}
\label{sec:data:subsec:analysis}

Subsequently, we analyze the dataset.
First, we present the subjects' characteristics and some metadata.
Among other things, we will show the distribution of native speakers, programming experience, and statistics on how much time the participants spent to complete the study.
In the second part, we analyze the dataset in terms of its linguistic features.
This includes the word count, the used vocabulary, and an in-depth consideration of frequently used wording and description structures.
Finally, we answer the research questions and discuss the validity of our approach.

\subsubsection{Subject Characteristics \& Metadata}

\begin{table}[t]
	\centering
	\caption{Distribution of native English speakers and participants with programming experience among the accepted, rejected, and all submissions.}
	\label{tab:dist_natives_programming}
	\begin{tabular}{lrrr}
		& accepted  & rejected & all \\
		\hline
		native speaker    & .60 	& .51	& .60 \\
		non-native sp.  	& .40 	& .49	& .40 \\
		\hline
		prog. skills      & .28	& .49	& .30 \\
		no prog. sk. 		& .72	& .51	& .70 \\
	\end{tabular}
\end{table}

\begin{table}[t]
	\centering
	\caption{The Distribution of the ten most frequent countries of origin (Alpha-3 country codes -- ISO 3166).}
	\label{tab:origin}
	\begin{tabularx}{\columnwidth}{XXXXXXXXXX}
		GBR & USA & POL & PRT & ESP & CAN & ITA & MEX & GRC & NLD \\
		\hline
		216 & 103 & 66 & 47 & 26 & 26 & 26 & 21 & 14 & 12
	\end{tabularx}
\end{table}

\begin{figure}
	\begin{center}
	\begin{tikzpicture}
	\begin{axis}[
	ylabel=Quantity,
	xlabel=Age,
	y post scale=0.53,
	ybar,
	ymin=0,
	ymax=46,
	/pgf/bar width=2pt,
	]
	\addplot[black,fill=black]
	coordinates{
		(22.0,45)
		(21.0,40)
		(20.0,38)
		(23.0,37)
		(24.0,36)
		(30.0,35)
		(26.0,35)
		(25.0,35)
		(18.0,35)
		(28.0  ,  35)
		(19.0  ,  34)
		(27.0  ,  26)
		(29.0  ,  25)
		(	31.0  ,  23)
		(	32.0  ,  23)
		(	33.0  ,  22)
		(	41.0  ,  20)
		(	38.0  ,  19)
		(	34.0  ,  19)
		(	36.0  ,  17)
		(	40.0  ,  16)
		(	37.0  ,  15)
		(	39.0 ,   13)
		(	35.0 ,   12)
		(	42.0  ,  12)
		(	48.0  ,  10)
		(	44.0  ,   8)
		(	47.0  ,   6)
		(	46.0  ,   5)
		(	50.0  ,   5)
		(	45.0  ,   5)
		(	56.0  ,   5)
		(	43.0  ,   5)
		(	53.0  ,   4)
		(	54.0  ,   4)
		(	51.0  ,   3)
		(	61.0  ,   3)
		(	55.0  ,   3)
		(	63.0  ,   2)
		(	59.0  ,   2)
		(	58.0  ,   2)
		(	49.0   ,  2)
		(	52.0  ,   2)
		(	68.0  ,   1)
		(	69.0  ,   1)
		(	65.0  ,   1)
		(	57.0  ,   1)
		(	60.0  ,   1)
		(	66.0  ,   1)
		(	75.0   ,  1)
		(	62.0   ,  1)
		(	76.0  ,   1)};
	\end{axis}
	\end{tikzpicture}
	\caption{Age distribution of the participants.}
	\label{fig:age}
\end{center}
	\end{figure}
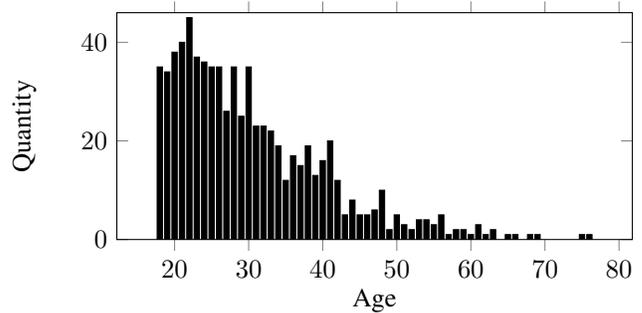

 \begin{table}[t]
	\begin{center}
		\caption{The Duration of the Participation per Subject (in Minutes).}
		\label{fig:duration}
		\begin{tabular}{lrr}
			& accepted & rejected \\
			\hline
			average & 9 & 17 \\
			minimum & 5 & 1 \\
			maximum & 25 & 25 \\
			standard derivation & 6 & 6 \\
		\end{tabular}
	\end{center}
\end{table}

The majority of the 870 participants are native English speakers (522, 60\%).
Regarding the proportion of accepted vs. rejected submission we note that more than half (154, 51\%) of the rejected were submitted by native speakers.
However, among the accepted submission we observe the 60 (native) to 40 (non-native) ratio again.
609 (70\%) of all participants have no programming experience.
This is an encouraging fact, since we want to investigate verbalizations of laypersons, not programming experts.
Interestingly, submissions by subjects without programming experience are more often accepted.
Among the rejected submission the ratio is again nearly half-half.
\autoref{tab:dist_natives_programming} summarizes the distributions.
This personal data was collected through the conclusive questionnaire of our study (see \autoref{sec:data:subsec:conduction}).
Besides, Prolific shares the personal data of all participants with the initiator of a study.
Note that this data is incomplete since users provide it voluntary (during registration).
The evaluation of this data revealed that females and males participated virtually equally (359 females and 366 males).
\autoref{fig:age} shows the distribution of the age of the participants.
The age of the youngest participants was 18, the oldest participant was 76; 
the mean age was 29.45.
However, \autoref{fig:age} clearly shows that the majority of the participants are 30 and younger (456, 59\%);
445 of the participants are students.
According to their own statements, 306 are fully employed, 150 part-time employed, 90 unemployed, and 132 chose \enquote{other} as employment status.
\autoref{tab:origin} depicts the ten most frequent countries of origin.
The majority of the participants was born either in Great Britain or the US.

Prolific also records the duration of each session.
We summarize the statistics in \autoref{fig:duration}.
Here we distinguish accepted and rejected submissions again.
The table shows that the range for the accepted submission is 5 to 25 minutes, with an average of 9 minutes.
For the rejected submission we can observe the same maximum\footnote{Note that we have set a timeout after 25 minutes (see \autoref{sec:data:subsec:conduction}).}.
However, the minimum is one minute only and the average is 17 minutes.
This suggests that participants that got rejected either worked rather sloppy or were  unable to cope with the task.

\subsubsection{Dataset Analysis}

 \begin{table}[t]
	\begin{center}
		\caption{Statistics on the number of words per submission in total.}
		\label{tab:length}
		\begin{tabular}{rrrrr}
			min & max & mean & st. dev. & sum (total)\\
			\hline
			1 & 312 & 35.43 & 22.48 & 109008
		\end{tabular}
	\end{center}
\end{table}
\begin{table}[t]
	\begin{center}
		\caption{The uniquely used words per user and per scenario.}
		\label{tab:words_per_user}
		\setlength{\tabcolsep}{3.9pt}
		\begin{tabular}{rrrr|rrrrr}
			\multicolumn{4}{c}{per user}&\multicolumn{5}{c}{per scenario}\\
			 & & & &sc. 1 & sc. 2 & sc. 3 & sc. 4 &  \\
			min & max & mean & st. dev. &(greet) & (coffee) & (drinks) & (table) & corpus \\
			\hline
			2 & 227 & 70.69 & 26.70 & 570 & 625 & 691 & 685 & 1469\\
		\end{tabular}
	\end{center}
\end{table}

 \begin{table}[t]
	\begin{center}
		\caption{Six exemplary submissions taken from different scenarios.}
		\label{tab:submissions}
		\begin{tabularx}{\columnwidth}{rrX}
			ID & Sc.  & Submitted Text \\
			\hline
			302 & 1 & Look directly at the person. Wave your hand. Say 'hello'.\\
			1000 & 2 & You have to place the cup under the dispenser and press the red button to make coffee.\\
			1346 & 2 & Making coffee means you have to press the red button, put a cup underneath the hole and then pouring the coffee that comes out into your cup\\
			2180 & 3 & To ring a beverage, open the fridge and select one of te beverages inside, pour it into one of the glasses on the kitchen counter and hand the glass over to the person.\\
			2511 & 4 & collect cutlery from cupboard, bring them to the table and place down neatly\\
			2577 & 4 & To set the table for two, Go to the cupboard and take two of each; plates, glasses, knives, and forks. Take them to the kitchen table and set two individual places.\\
		\end{tabularx}
	\end{center}
\end{table}

The dataset contains 3168 submissions for four scenarios with nearly 110,000 words in sum.
\autoref{tab:length} shows statistics on the number of words.
The number of words per solution ranges from one to 312 with a mean of 35.43.
The standard derivation is 22.48.
The wide range and  relatively high standard derivation suggests that the subjects made use of the possibility to verbalize their intents freely.

This assumption is also supported by the vocabulary the participants used as depicted in \autoref{tab:words_per_user}\footnote{We lemmatized all submissions to determine the unique words.}.
The range of unique words each participant used is two to 227.
On average, the subjects used more than 70 words.

\autoref{tab:words_per_user} additionally shows the vocabulary per scenario and for the entire dataset.
The number of unique words per scenario is almost the same for all scenarios.
The subjects used about 600 words for each scenario (min: 570, scenario 1, max: 691 scenario 3).
The total number of unique words is 1469.

\begin{figure}[t]
\begin{center}
	\begin{tikzpicture}
	\begin{axis}[
	ylabel=Quantity,
	x tick label style={rotate=90,anchor=east,},
	y post scale=0.82,
	ybar,
	ymin=0,
	ymax=15400,
	xmin=the,
	enlarge x limits=0.02,
	nodes near coords,
	every node near coord/.append style={font=\tiny,rotate=90, anchor=west},
	x post scale=1.1,
	xtick=data,
	/pgf/bar width=2pt,
	every axis y label/.style={
		at={(ticklabel* cs:0.5,15)},
		anchor=south,
		rotate=90
	},
	symbolic x coords={
		the,
		to,	
		and,
		you,
		glass,
		two,
		be,
		table,
		it,
		beverage,
		in,
		coffee,
		kitchen,
		machine,
		then,
		cup,
		hand,
		plate,
		on,
		place,	
		take,
		from,
		fridge,
		of,
		bring,
	},
	]
	\addplot[black,fill=black]
	coordinates{
		(the,12662)
		(to,7257) 	
		(and,4728)
		(you,4062)
		(glass,	2456)
		(two,	2384)
		(be,	1712 	)
		(table,	1706)
		(it,	1678 	)
		(beverage,	1670 	)
		(in,	1611 )
		(coffee, 	1518 	)
		(kitchen, 	1418 )
		(machine, 	1402 )
		(then, 	1385 	)
		(cup,	1372 	)
		(hand,1249 	)
		(plate,1189 )	
		(on,1134 	)
		(place,1120 )	
		(take,1091 	)
		(from,1083 	)
		(fridge,	1039 )	
		(of,1027 	)
		(bring,1025)
	};
	\end{axis}
	\end{tikzpicture}%
	\caption{Distribution of the 25 most frequent words in the dataset.}
	\label{fig:most_frequent_words}
\end{center}%
\end{figure}
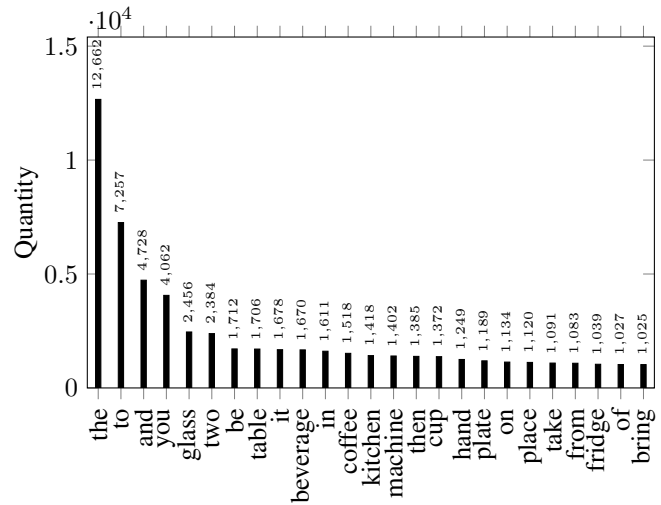%

The 25 most frequently used words are depicted in \autoref{fig:most_frequent_words}.
Most of them are stop words; the word \emph{the} alone accounts for 11.6\% of the usages.
The four most frequent words (\emph{the}, \emph{to}, \emph{and}, and \emph{you}) make up more than 26.3\%.
Considering stop words in our context is reasonable, since some words commonly declared as stop words are valuable to identify \emph{specifying} or \emph{declarative} parts in descriptions, e.g. \emph{to} in phrases like \emph{to prepare coffee}.
The first word that is not a stop word is \emph{glass} at rank five.
The word frequencies in our dataset comply with Zipf's law~\cite{zipfSelected1932}\footnote{Basically, Zipf's law says that the word frequency of any given word within a natural language corpus is inversely proportional to its rank.}.

To gain a deeper understanding of how the subjects verbalized new functions, we have analyzed all submissions in terms of content and linguistic structure.
\autoref{tab:submissions} shows six typical submissions.
These examples show that submissions contain typos (\enquote{ring some beverage [...] te beverages inside} in 2180), grammar mistakes (\enquote{pour beverage to the glass}), and are of differing styles (full sentences vs. notes).
Moreover, we observed that the narrative perspective varies.
Mostly, the subjects acted in the role of a teacher, sometimes as a developer.
Some subjects shifted their perspective to a naive end-user.
Pane et al. observed similar behaviors in their study.
Most response are in present tense, using an imperative mood.
However, we also found many declarative sentence and even interrogative clauses.
Less frequently, responses contained passive voice or conjunctive mood.

The most notable observation in terms of the studies objective: more than a third of the submissions (1170, 37\%) do not contain a description of new functionality (see \autoref{fig:stats}). 
Instead, they are merely a sequence of actions.
Submission 302 and 2511 are typical examples.
The large number of this kind of submissions may be attributed to a rather restrictive labeling of ours. 
However, since the objective of our study was to analyze unrestricted language for function descriptions, we searched for any hint that a submission comprises a teaching intent.
Thus, we have labeled submissions such as 1000 as function description, even if only the word \emph{to} in the phrase \emph{to make coffee} indicates an eventual function declaration.

 \begin{table}[t]
	\caption{Distribution of the binary and ternary labels.}
	\label{fig:stats}
	\begin{center}
		\setlength{\tabcolsep}{2.6pt}
		\begin{tabular}{rrr|rrrr}
			\multicolumn{3}{c|}{binary} & \multicolumn{4}{c}{ternary}\\
			func & non-func & sum & decl & spec & misc & sum\\
			\hline
			1998 (.63) & 1170 (.37) & 3168 & 15559 (.21) & 57156 (.76) & 2219 (.03)& 74934
		\end{tabular}
	\end{center}
\end{table}

Regarding the general structure of the submission with explicit function description we found many similar contributions.
Often, the participants start with a brief greeting, like \enquote{Hi Armar [...]} or \enquote{Hey robot [...]}, proceed with the function declaration, and finish with the specification of intermediate steps.
The position of the declarative part differs strongly.
Many participants also put it at the end (or place it anywhere).
A notable share of participants also puts the declaration before the specification but repeats it afterwards, often with a different wording or even on another level of abstraction.
A few submissions lack a specification of intermediate steps.

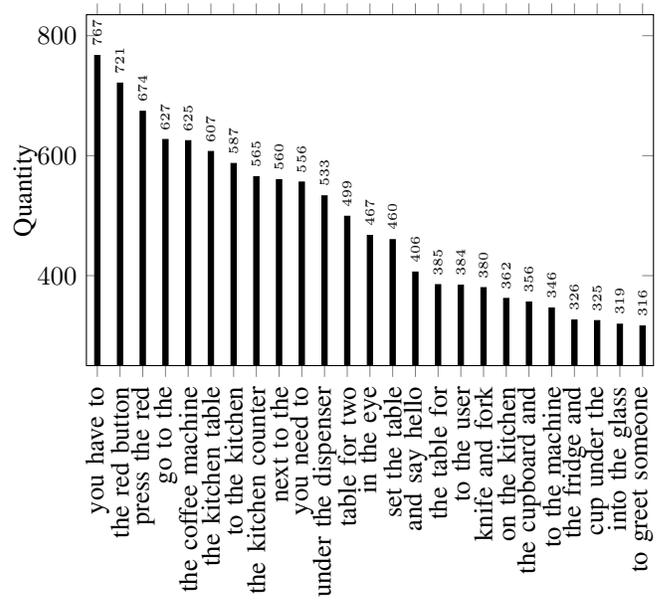
\begin{figure}[t]
	\begin{center}
		\begin{tikzpicture}
		\begin{axis}[
		ylabel=Quantity,
		x tick label style={rotate=90,anchor=east,},
		y post scale=0.82,
		ybar,
		ymin=250,
		ymax=835,
		xmin=you have to,
		enlarge x limits=0.02,
		nodes near coords,
		every node near coord/.append style={font=\tiny,rotate=90, anchor=west},
		x post scale=1.1,
		xtick=data,
		/pgf/bar width=2pt,
		every axis y label/.style={
			at={(ticklabel* cs:0.5,15)},
			anchor=south,
			rotate=90
		},
		symbolic x coords={
			you have to,
			the red button,
			press the red,
			go to the,
			the coffee machine,
			the kitchen table,
			to the kitchen,
			the kitchen counter,
			next to the,
			you need to,
			under the dispenser,
			table for two,
			in the eye,
			set the table,
			and say hello,
			the table for,
			to the user,
			knife and fork,
			on the kitchen,
			the cupboard and,
			to the machine,
			the fridge and,
			cup under the,
			into the glass,
			to greet someone,
		},
		]
		\addplot[black,fill=black]
		coordinates{
			(you have to,767)
			(the red button,721)
			(press the red,674)
			(go to the, 	627)
			(the coffee machine, 	625)
			(the kitchen table, 	607)
			(to the kitchen, 	587)
			(the kitchen counter, 	565)
			(next to the, 	560)
			(you need to, 	556)
			(under the dispenser, 	533)
			(table for two, 	499)
			(in the eye, 	467)
			(set the table, 	460)
			(and say hello, 	406)
			(the table for, 	385)
			(to the user, 	384)
			(knife and fork, 	380)
			(on the kitchen, 	362)
			(the cupboard and, 	356)
			(to the machine, 	346)
			(the fridge and, 	326)
			(cup under the, 	325)
			(into the glass, 	319)
			(to greet someone,316)};
		\end{axis}
		\end{tikzpicture}
		\caption{Distribution of the 25 most frequent trigrams in the dataset.}
		\label{fig:most_frequent_3grams}
	\end{center}
\end{figure}

Although some wordings occur more frequently, the submissions differ strongly.
The syntax is highly different even for the same scenario.
Due to variability of the English language, e.g. different sentence construction or the use of synonyms, we received strongly varied descriptions.
However, we observed that many subjects used gerunds (e.g. \enquote{preparing coffee}) or to-infinitives (e.g. \enquote{to prepare coffee}) in combination with certain words or phrases (e.g. \enquote{means} or \enquote{you have to}) to express a teaching intent.
Moreover, we found the subjects tend to use certain sentence structures and wordings to verbalize their wish to extend the functionality, such as \enquote{we are going to learn how to [...]}, \enquote{in order to [...] you have to [...]}, \enquote{if you want to [...] you need to [...]}, or \enquote{[...] means you have to [...]} (see submission 1346).
To further investigate this observation, we have extracted the most frequent trigrams from the dataset (see \autoref{fig:most_frequent_3grams})\footnote{We have extracted the most frequent n-grams for n=[2;4]. However, the trigrams are most informative}.
Besides many domain specific phrases, one will notice the trigrams \emph{you have to} and \emph{you need to}.
The first is the most frequent trigram, the latter at rank nine.
Both are used to separate the declaration from the specification (like in example E1 in \autoref{fig:example-labels}).
In 1327 of the submissions one of these two phrases was used (variations not included).
That supports the assumption that subjects use certain wordings for this task.
The distribution of the ternary labels that depict the distinct parts of a submission with teaching intent is shown in \autoref{fig:stats}.
The majority of words were used to specify the intermediate steps (.76);
words used to verbalize the teaching intent, including the wish for extension and the name for the new functionality, account for more than a fifth (.21).
Interestingly, only a negligible number of words are not useful in our context (.03). 

\subsubsection{Research Questions}

With the dataset at hand we can answer the research questions (see \autoref{sec:data:subsec:design}).
The analysis of the data revealed that more than a third of the submissions do not contain a function description.
Therefore, we conclude that subjects \emph{not} always clearly state that they wish to add new functionality (\emph{RQ1}).

For the submissions that contained a function description we were able to label the parts, which contain the declaration of the function.
Even though the declarative part is often distributed over the submission, it can be clearly identified in almost all cases (\emph{RQ2}).

During the analysis of the dataset 
we observed that many submissions contain phrases that do not carry useful information in our context (\emph{RQ3}).
However, these phrases account for 3\% of the words in our dataset only.
Additionally, they can be clearly separated from the informative sections.

Finally, we were able to discover two trigrams subjects repeatedly used to state their wish to teach a new function.
However, even though we found some tendencies in the general structure of the function description, we were not able to find patterns that are valid for the majority of the submissions.
Thus, we can not give a definite answer to \emph{RQ4}.

\subsubsection{Validity}

We briefly discuss the validity of our approach. 
The rejection of 8.78\% of the submission is arguably a sort of experimenter effect and threatens the internal validity.
However, Christensen argues, that this approach is reasonable, since this kind of study aims to analyze the participants verbal knowledge transfer and not their willingness to achieve good results or their interpretation of the exercises~\cite{christensenExperimental2007}.
Another issue is the difference between spoken and written language.
Originally, we aimed to study how laypersons actually talk to robots.
However, for reasons we discussed in \autoref{sec:data:subsec:design} we discarded the idea of recording audio files.
Instead, we encouraged the subjects to respond spontaneous and imagine that they talk to the robot.
However, we can not be sure whether our findings apply to \enquote{real} spoken input.
Finally, all scenarios comprise the same actor and the same environment, which is an issue regarding the generality of the dataset.
Of course, our findings do not generalize; subjects might act differently in other domains.
However, we believe that some of our findings are generally valid since they do not depend on the domain, e.g. the prevalent usage of the imperative mood to describe intermediate steps, similar wordings to express the teaching intent, or frequent 3-gramms to separate the declaration from the specification.

\section{Conclusion \& Future Work}
\label{sec:conclusion}

We have presented an empirical study on how laypersons teach robots new functions.
With the study we aimed at examining the verbalization
humans use in this situation.
We used the micro-tasking platform Prolific to collect data from a broad spectrum of participants.
In four scenarios subjects were supposed to teach a robot new skills using nothing but natural language.
The scenarios all featured the household robot ARMAR-III in a kitchen environment but strongly differing skills to teach, e.g. \emph{preparing coffee} or \emph{greeting someone}.

We were able to collect 3618 descriptions from 870 participants.
The analysis of the dataset revealed that more than one third of the submissions do not contain a teaching intent.
Instead, the subjects solely verbalized a sequence of actions, i.e. the intermediate steps, in most of these cases.
For the submission that comprised a function description, we were able to separate the function declaration (composed of wish for extension and function name), the description of intermediate steps, and irrelevant phrases.
However, we observed that participants tend to mix these parts.
To record these results we labeled all submissions.
A binary label set indicates whether a function description is (non-)existent.
Additionally, we label the aforementioned parts on a per-word level.
We observed that the participants use certain phrase to express their wish to teach a new function, such as \emph{you have to}.
Two of these phrases are among the most frequent trigrams in our dataset.

With the corpus we aim to foster the research in the area of programming with natural language and in particular the teaching of functionality.
We will further investigate the linguistic characteristics of verbalized function descriptions, preferably in collaboration with the community.
Another logical next step is the creation of classifiers for the proposed label sets.
We will examine whether classical machine learning approaches or more advanced approaches 
are more appropriate. 
In the long term, we plan to automatically synthesize method definitions from such natural language descriptions.
We hope not only to generate valid method names and parameters, i.e. signatures, but also to synthesize the method body.


\bibliographystyle{IEEEtran}
\bibliography{icsc2020_command_extraction}

%
\IEEEpeerreviewmaketitle

\end{document}